\documentclass[9pt]{IEEEtran}
\pdfoutput=1
\usepackage{fancyvrb}
\fvset{numbers=left,fontsize=\footnotesize}
\DefineShortVerb{\|}
\usepackage{paralist}
\usepackage{graphicx}

\title{A cookbook of translating English to Xapi}
\author{
\IEEEauthorblockN{Ladislau B{\"o}l{\"o}ni}\\
\IEEEauthorblockA{
Dept. of Electrical Engineering and Computer Science\\
University of Central Florida\\
Orlando, FL 32816--2450\\
lboloni@eecs.ucf.edu
}
}
\IEEEspecialpapernotice{(XTR-001)}

\begin{document}
\maketitle

\begin{abstract}
\begin{quote}

The Xapagy cognitive architecture had been designed to perform {\em narrative reasoning}: to model and mimic the activities performed by humans when witnessing, reading, recalling, narrating and talking about stories. Xapagy communicates with the outside world using Xapi, a simplified, ``pidgin'' language which is strongly tied to the internal representation model (instances, scenes and verb instances) and reasoning techniques (shadows and headless shadows). While not fully a semantic equivalent of natural language, Xapi can represent a wide range of complex stories. We illustrate the representation technique used in Xapi through examples taken from folk physics, folk psychology as well as some more unusual literary examples. We argue that while the Xapi model represents a conceptual shift from the English representation, the mapping is logical and consistent, and a trained knowledge engineer can translate between English and Xapi at near-native speed. 

\end{quote}
\end{abstract}

\section{Introduction}
\label{sec:Introduction}

Xapagy is a cognitive architecture focusing on narrative reasoning. The objective of the system is to mimic the human behavior with regards to reasoning about stories. The reasoning in Xapagy is based on {\em autobiographical memory} - newly encountered events or story snippets are {\em shadowed} by elements of the memory, while {\em headless shadows} project into the future, providing prediction or in the past, providing detection of elements which might be missing in the current narrative.

The interaction of the Xapagy agent with the outside world happens through the Xapi {\em pidgin} language, a language with a simplified syntax, which approximates closely the internal representational structures of Xapagy. In the following we will provide a brief primer on Xapi and the internal structures it represents. 


A Xapi story is a sequence of sentences. Sentences can be in the simple subject-verb-object, subject-verb or subject-verb-adjective form. A single more complex sentence exists, in the form of subject-communication verb-scene-quote, where the quote is an arbitrary sentence which is evaluated in a different scene. Subjects and objects are represented as {\em instances} which can acquire various attributes in form of {\em concepts}. In general, each sentence corresponds to an internal Xapagy structure called a {\em verb instance}. Verb instances are stored in the autobiographical memory in a raw form. They can, however, be linked with succession, summarization, context and coincidence {\em links}.

As we shall see, one of the unexpected features of Xapagy instances is that they do not exactly correspond to entities in human speech. An entity in colloquial speech is often represented with more than one instance in Xapagy. These instances are often connected with {\em identity relations} but participate independently in verb instances, shadows and headless shadows.

Xapagy uses autobiography to perform reasoning, it can not reason on logical basis or first principles. In conclusion, the quality of the reasoning of an agent depends on the quality and representativeness of its autobiography, which includes direct and indirect experiences, stories which had been witnessed, heard or read. As we have a relatively large set of stories available in narrative form, translating stories from English to Xapi is an important step in the construction of a synthetic autobiography. 

Xapi, as a ``pidgin'' language, uses an English vocabulary. Stories described as simply a series of consecutive actions are trivial to translate. Let us assume the following snippet placed in the world of the Iliad:

\begin{quote}
{\em
Hector is a Trojan warrior. Achilles is a Greek warrior. Achilles has a shield. Achilles's shield is big. Achilles hits Hector. Hector strikes Achilles.
}
\end{quote}

This can be translated into Xapi as follows:
 
\begin{quote}
\begin{Verbatim}
A "Hector" / exists.
"Hector" / is-a / trojan warrior.
An "Achilles" / exists.
"Achilles" / is-a / greek warrior.
"Achilles" / owns / a shield.
The shield -- of -- "Achilles" / is-a / big.
"Achilles" / hits / "Hector".
"Hector" / hits / "Achilles". 
\end{Verbatim}
\end{quote}

There are a number of situations where English to Xapi translation is not possible. One such case is self-referential English sentences such as {\em ``This sentence has five words''}. Similarly, sentences where the reference is made on the expressive form unlikely to be fully semantically translatable. For instance, the statement {\em The song ``Love ...  thy will be done'' by Prince is a religious parable applied to love.} is self-evident in English, due to the archaic English form chosen in the title. Naturally, such subtleties are not present in Xapi. In general, everything which is difficult to translate, let us say from English to French, would be near impossible to translate from English to Xapi. 

The objective of this report is to help human operators translate English into Xapi by offering a cookbook of translation solutions for specific situations. As a note, for reasons of conciseness, the Xapi snippets will sometimes be abbreviated to the most interesting parts. The full Xapi stories for the specific translations are downloadable form the Xapagy webpage.

The remainder of this report is organized as follows. Section~\ref{sec:NeedChallenge} discusses trivial cases of representation and identify the nature of challenges which are encountered when translating more complex stories. 

Complex semantic structures are represented in English using complex {\em syntax}. Xapi sentences have a very simple syntax, thus the semantic of the story must be expressed through other tools, in particular by scenes and scene relations, identity relations between instances and the notion of change and coincidence links. This is covered in Section~\ref{sec:RepresentationTools}. 

Section~\ref{sec:FolkPhysics} discusses specific solutions for representational challenges covering the area of folk physics, while Section~\ref{sec:FolkPsychology} deals with examples drawn from folk psychology. 

As a note, the Xapagy system is actively developed and not all the representational challenges are solved. In this report we will describe future work is separate notes using italic text.

%
%
\section{Representation tools in Xapi}
\label{sec:RepresentationTools}

We have seen that translation is trivially possible when we have simple sequences of consecutive actions and we have seen that there are cases where fully semantic translation is unlikely to be possible. In between these two cases, there is the wide range of cases where fully semantic translation is possible, but it requires a translation from the grammatical concepts to the way Xapagy thinks about things. 

The reader will find the Xapagy internal structures somewhat counter-intuitive, and indeed, the shift from the text to the Xapagy internal representation (and Xapi) requires a certain non-trivial shift of perspective. This is similar to the way in which a linguist considers a sentence in terms of its syntax tree - a representation which is counter-intuitive for non-specialists.

In the following we will discuss four tools of the Xapagy / Xapi representational model which allow us to represent significantly more stories than the trivial ones we discussed before. These are the (a) scenes and scene relations (b) identity relations (c) change and (d) coincidence.

%
%
\subsection{Scenes and scene relations}

Human languages do not limit themselves to the enumeration of events witnessed in the current moment. They deal with the past, future, hypothetical situations, plans and so on. These are represented in language in form of categories such as tense, modality, voice and so on. 

Such situations are expressed in Xapagy by placing the actions and the instances which participate in them in different {\em scenes}. Instances in different scenes are often connected through identity relations. The grammatical category is then represented using the relationship between the scenes. 

The most unusual part of this representation is the fact that we have instances replicated in the different scenes. 

If we consider a situation when Achilles remembers his fight with Hector, the common sense interpretation tells us that we are talking about a single Greek warrior. The Xapagy representation, however, involves {\em two} instances of Hector in different scenes, with possibly different attributes and, naturally, with different shadows. 

In conclusion, the task of the translator when encountering a specific grammatical category in English is:

\begin{compactitem}
\item[-] identify what instance in what scene it refers to 
\item[-] if necessary, create a new scene, new instance and set up the appropriate identity relations. 
\end{compactitem}

%
%
\subsection{Identity relations - further uses}

We have seen the use of identity relations to express grammatical categories such as tense and modality. This can be (in theory) automatically generated using {\em syntactic parsing}. For instance, if we determine that some parts of a text are in past tense, we put it in a past scene. 

Identity relations between instances are used in some less obvious cases as well, where the scenes represent situations which in the English text are encoded in the semantics, not in the syntax.

Let us consider the famous dialog between Little Red Riding Hood and the wolf dressed up as Grandma. There are two planes of interpretation: one involves a grandma-granddaughter conversation, while the other one is a predator-prey interaction. These have naturally different interpretations, thus they need to be shadowed differently. The Xapagy representation, as expected, involves two scenes: one involving LRRH with Grandma, while the other one LRRH and the wolf.

\if defined
, with the identity relations set up as in Figure~\ref{fig:LRRH}.

\begin{figure*}
\begin{center}
\includegraphics[width=\columnwidth]{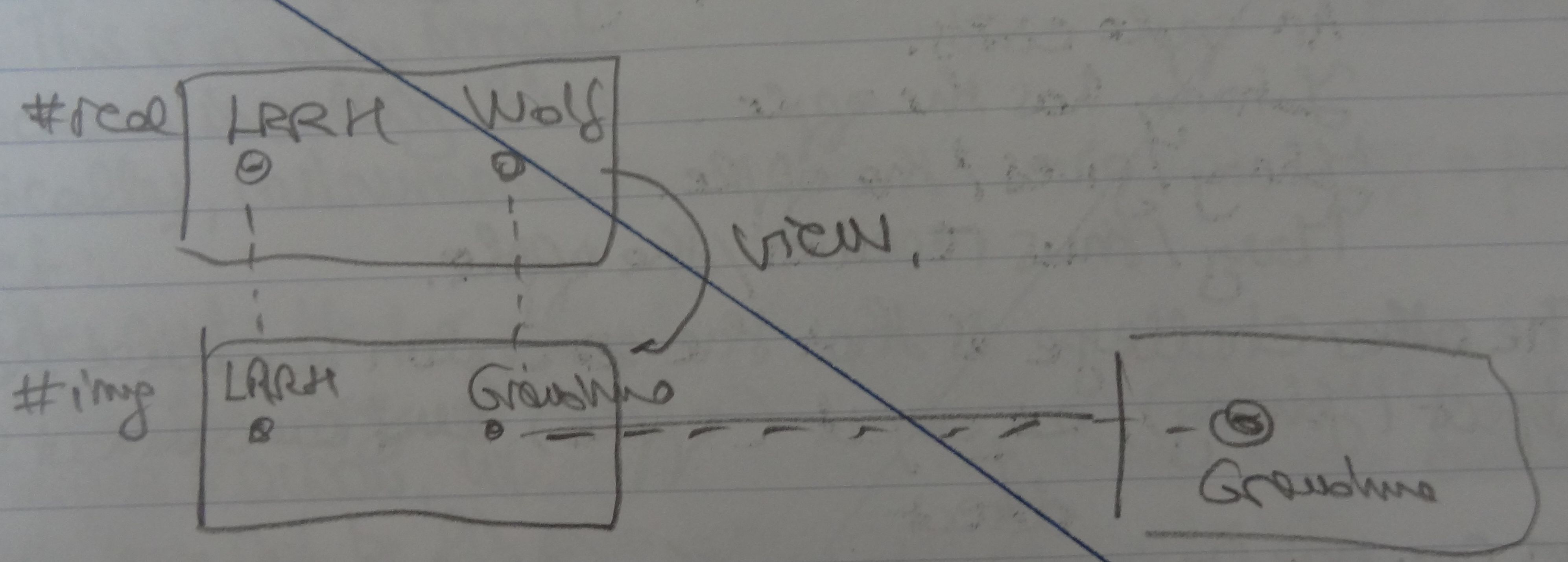} 
\caption{\label{fig:LRRH} Using the identity relations to describe the conversation of LRRH with the wolf disguised as Grandma.
}
\end{center}
\end{figure*} 
\fi

From the point of view of the human translator, she must identify situations when a certain story snippet involves multiple views and represent them with the appropriate number of scenes and identity relations. Such situations often involve cases where the different instances have different attributes (sometimes, as in the case of the wolf disguised as Grandma, radically so).

It is also possible that one of the instances is further connected with an identity relationship to somewhere else, but not the other. In our case, Grandma imagined by LRRH is connected to the real Grandma from other scenes, but not the wolf.

\if defined
{\em {\bf Future work:} The fact that the scene is imaginare might be inferrable from the quote-verbs used. We can say LRRH / imagines in #img // Wolf / is-identical / Grandma -- in -- scene #x. 

This would require for the VIs to be treated differently when they appear as a quote, this is not currently done in Xapagy.

}
\fi

%
%
\subsection{Change}

The Xapagy framework posits that the acquisition of new attributes, for an instance is an act of {\em discovery} not {\em change}. If we say:

\begin{quote}
\begin{Verbatim}
"Hector" / is-a / trojan.
\end{Verbatim}
\end{quote}

\noindent it does not mean that he had just became a Trojan, but that he was a Trojan all the time and we have just discovered it. In conclusion, the attributes which can be added with |is-a| must be cummulative, they must be compatible with the previous attributes. Naturally, however, the new attributes added through |is-a| will affect the shadows and headless shadows in the future. 

What colloquial speech perceives as a change in the attributes of an entity, in Xapagy is represented by a verb which creates a separate new instance, which is identity-related to the old instance, which carries over some of the attributes of the old instance but also has new, possibly incompatible attributes:

\begin{quote}
\begin{Verbatim}
"Hector" / changes / dead.
\end{Verbatim}
\end{quote}

The intuition behind this approach is that the two instances will have radically different shadows - the live Hector will be shadowed by other warriors, other Trojans and so on, while the dead Hector will be shadowed by other corpses. Based on this, the Xapagy system can make appropriate predictions of what Hector will do (which are obviously different for Hector the warrior and Hector the corpse).

%
%
\subsection{Coincidence links}

Coincidence links connect multiple VIs together. The intuition behind coincidence links is that these VIs describe different {\em aspects} of a certain real world event. Sometimes coincidence is used to describe different participants in an event (when this cannot be done using only a subject and object). In other cases, it is used to represent the relation between a physical event and its significance in another plane of thinking (which would be likely represented in another scene). Coincidence links are created by adding the |thus| verb to the sentences which must be connected to a previous sentence, using labels to resolve ambiguous cases.

\section{Folk physics}
\label{sec:FolkPhysics}

This section contains examples of representation situations which we can label as the general term of folk physics. This includes things such as instrumentality, ownership, part relations and number sense.

%
%

\subsection{Ownership exchange}

Ownership in Xapagy is represented with a relation between the owner and the owned entity. The same concept applies to groups of relations such as physical containment, physical attachment, legal ownership and so on. The relationship is considered {\em active} as long as the relation VI is in focus. A relation VI can be expired from the focus, and a similar relation added later.

Let us now discuss how we can represent the exchange of ownership, for instance in the following snippet:

\begin{quote}{\em
Johnny gives an apple to Mary.
}
\end{quote}

This implies that there was an apple, which is in the ownership of Johnny. This is not an action, only a situation:

\begin{quote}
\begin{Verbatim}
An apple / exists.
"Johnny" / has / the apple.
"Johnny" / gives / an apple.
"Mary" / thus receives / the apple.
\end{Verbatim}
\end{quote}

The other challenge is that the English sentence has a subject (Johnny), a direct object (apple) and an indirect object (Mary). This can not be represented with a single VI, which at most allows a subject and an object. 

\begin{figure}
\begin{center}
\includegraphics[width=\columnwidth]{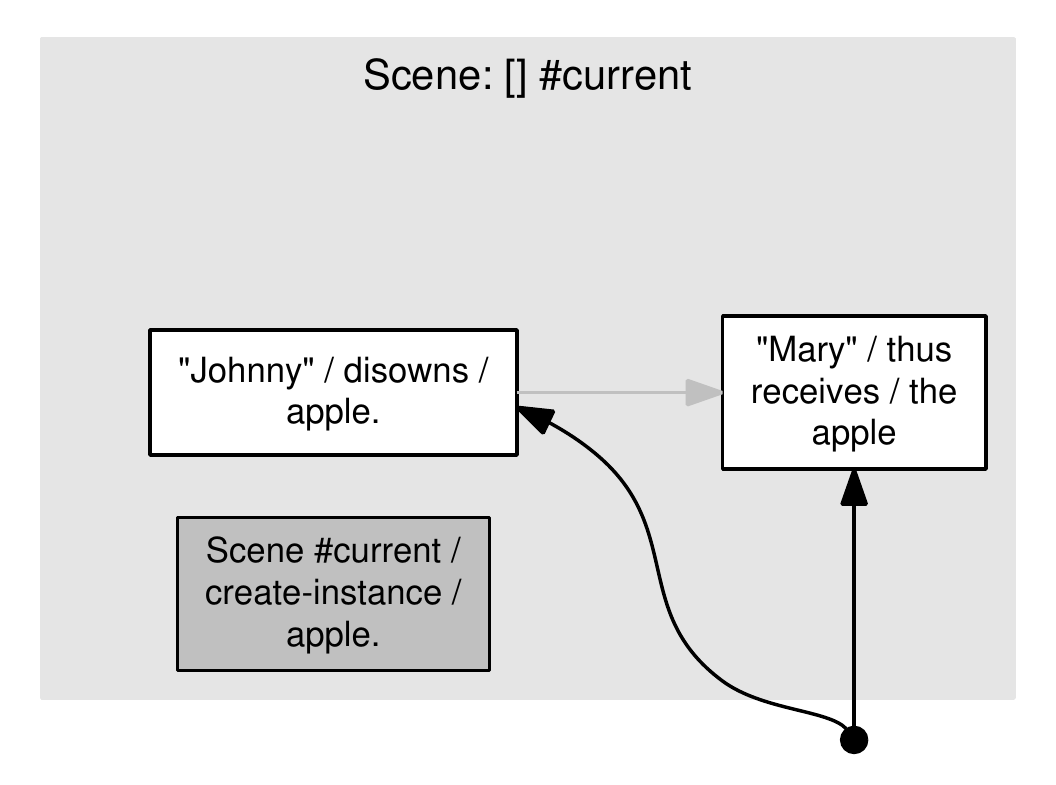} 
\caption{\label{fig:ComplexPhrases-901-FVI} Focus VIs for representing ownership exchange.
}
\end{center}
\end{figure}

The representation model splits the English statement into two VIs, one for Johnny giving the apple and the other one for Mary receiving it, but connects them with a coincidence link, which shows that these two VIs describe different aspects of a single event. Figure~\ref{fig:ComplexPhrases-901-FVI} shows the focus VIs for this exchange. 

One of the consequences of this representation model is that we can represent situations where we know only one half of the statement. For instance, we can say:

\begin{quote}{\em
Mary is given an apple.
}
\end{quote}

which is translated:

\begin{quote}
\begin{Verbatim}
An apple / exists.
"Mary" / receives / the apple.
\end{Verbatim}
\end{quote}

The second VI will be shadowed by sentences of the given form, some of them having coincidence links to the VI pair which specifies that there is somebody who gave the apple. In the course of the normal operation of the Xapagy agent, this will appear as headless shadow suggesting a missing or predicted action - depending on other settings, the agent might or might not act on these suggestions. 

%
%
\subsection{Instrumentality}

Instrumentality is the situation where an agent uses an external entity as an instrument to accomplish an action. This is a very important component of many conscious actions involving humans. Linguistically, instrumentality is shown in English using a prepositional object:

\begin{quote}{\em
Achilles cuts Hector with the sword.
}
\end{quote}

The challenge in translating this to Xapi is that this is clearly a single action, which, however, involves three entities which can not be fit into a single VI:

In Xapagy instrumentality is represented by an S-V-O VI using the verb |vm_Uses| with the subject being the agent which uses the instrument and the object being the instrument used. This verb is often part of a coincidence group which describes the complex action:

\begin{quote}
\begin{Verbatim}
"Achilles" / cuts / "Hector". 
"Achilles" / thus uses / the sword.
\end{Verbatim}
\end{quote}

\begin{figure}
\begin{center}
\includegraphics[width=\columnwidth]{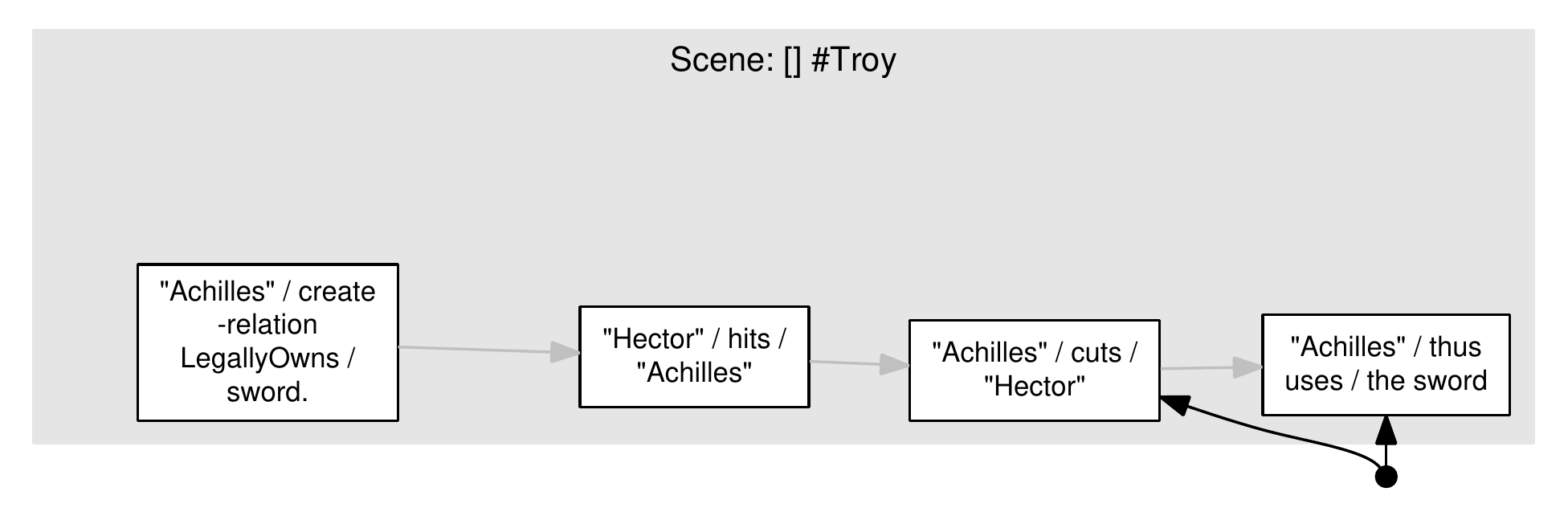} 
\caption{\label{fig:ComplexPhrases-501-FVI} Instrumentality represented as a coincidence group and the ``uses'' verb.
}
\end{center}
\end{figure}

The resulting focus VIs are shown in Figure~\ref{fig:ComplexPhrases-501-FVI}.

%
%
\subsection{Agency and achievement}

A frequently used language construct designates an entity as being the agency who brought about a certain state of the world, without explicitly describing the actions it used to do so. 

Narratives involving humans use this construct frequently. At the linguistical level, such constructs usually use the same very general verb (gets, puts, makes and so on) although in some cases, we have more specific verbs which depend on the achieved state (e.g. kills, destroys, repairs and so on).

\begin{quote}{\em
Achilles kills Hector.
}
\end{quote}

In Xapi, this representation is achieved by a coincidence group in which the subject agent is the subject of a verb |achieves|. 

\begin{quote}
\begin{Verbatim}
"Achilles" / achieves. 
"Hector" / thus changes / dead.
\end{Verbatim}
\end{quote}

\begin{figure}
\begin{center}
\includegraphics[width=\columnwidth]{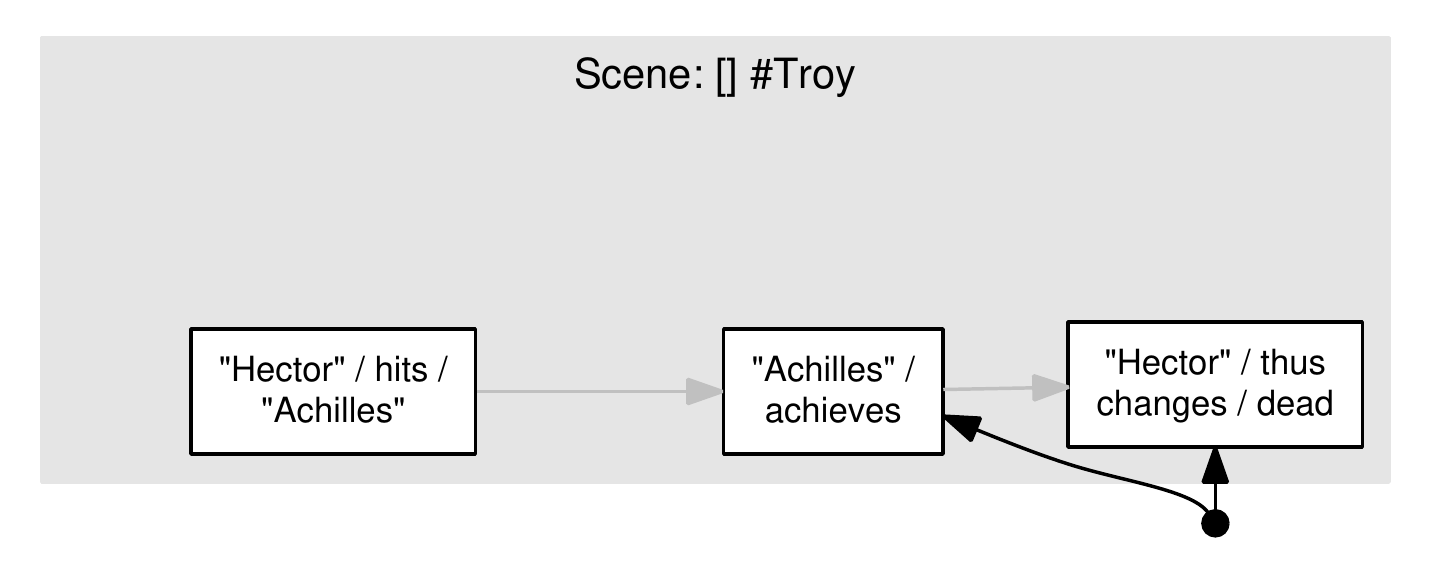} 
\caption{\label{fig:ComplexPhrases-502-FVI} Achievement represented as a coincidence group and the ''achieves'' verb.
}
\end{center}
\end{figure}

%
%
\subsection{Parts and lack of them}

An important aspect of folk physics is the reasoning about parts / whole relationships. These are represented by relationships in Xapagy, the basic part relationship being provided by the verb |contains|. 

We can say that Billy has a head by:

\begin{quote}
\begin{Verbatim}
A "Billy" / is-a / man.
"Billy" / contains / a head.
\end{Verbatim}
\end{quote}

The fact that most humans have a head is part of the autobiographical information of the agent. This needs to be put in contrast to default logic or frame systems, where there is a logical description of the structure.

In most stories, we do not need to enumerate the body parts of the agents. With a different autobiography, the Xapagy agent will maintain them as a headless shadow, thus when we refer to |the head -- of -- Billy|, the instance will be generated automatically.

{\em {\bf Future work:} this create on demand by reference is present in Xapagy but it is largely untested and used very rarely. It should get more use in larger autobiographies.}

The next question is how do we represent the fact that a certain part is missing, despite the fact that it would be predicted by the autobiography? The fact that it had not been enumerated in the current scene is not sufficient to indicate its absence.

Let us consider that we want to talk about the headless horseman. We need to represent the absence of the head {\em explicitly}. Simply not mentioning the existence of the head is not enough, as we have seen, for most humans the stories do not explicitly mention the existence of the head, yet we usually assume its presence.

This is accomplished by explicitly creating an instance for the head which has the |inexistent| attribute. This instance will be put in the appropriate part relation with the human. This would allow it to be explicitly shadowed by other headless humans from the autobiography:

\begin{quote}
\begin{Verbatim}
A "HeadlessHorseman" / is-a / man.
"HeadlessHorseman" / contains / an inexistent head.
\end{Verbatim}
\end{quote}

%
%
\subsection{Number sense: none, one, pair and many}

The Xapagy agent has a simple number sense which allows the representation of four classes of numbers: none, one, pair and many. In the following we will describe the semantics and representation of these concepts. 

\noindent {\bf None:} represents the absence of instances of a certain kind of objects in a certain context. For instance, we might say that there are no pencils in a box. This is represented through an |inexistent| instance as seen in the previous example. 

\begin{quote}
\begin{Verbatim}
A box / exists.
The box / contains / an inexistent pencil.
\end{Verbatim}
\end{quote}

\noindent {\bf One:} represents the existence of exactly one instance of a certain kind in a certain situation. For instance, we might say that there is one pencil in the box. This is represented as an instance directly mapping to the object. 

\begin{quote}
\begin{Verbatim}
A box / exists.
The box / contains / a pencil.
\end{Verbatim}
\end{quote}

\noindent {\bf Pair:} represents the existence of exactly two instances in a certain situation. This is represented through a group instance, which also has the |pair| attribute, with the two instances being members of the group.  

\begin{quote}
\begin{Verbatim}
A box / exists.
The box / contains / a pair pencil.
A pencil red / becomes-right-of / pair. 
A pencil blue / becomes-left-of / pair.
\end{Verbatim}
\end{quote}

\noindent {\bf Many:} represents a large number of instances in a certain situation. This is represented through the a group instance which also has the |many| attribute. The semantics of the many groups is that they are not enumerated - it is generally assumed that not all the members are individually 

\begin{quote}
\begin{Verbatim}
A box / exists.
The box / contains / a many pencil.
The many pencil / contains / a red pencil. 
\end{Verbatim}
\end{quote}

%
%
\subsection{Noun phrase modifiers and identity over time}

Most noun phrases are centered in Xapagy around a single instance. However, there are cases where the Xapi representation is counter-intuitive, as the specification provided by the modifier phrase takes place in a different scene:

\begin{quote}
{\em
The apple which was picked by John yesterday is red. 
}
\end{quote}

Commonsense interpretation would posit the existence of a single apple. In Xapagy, however, we are talking about two scenes: one of them the present reality, while the other the scene of yesterday. As every instance is member of a single scene, the implication is that we have two instances of the apple, connected with an identity relation:

\begin{quote}
\begin{Verbatim}
$NewSceneOnly #current, none, apple 
$NewSceneCurrent #yesterday, none, 
   apple-> apple, "John"
Scene #yesterday / is-past-of / scene #current.
Scene #yesterday / is-yesterday-of / 
   scene "current".
John / picks / the apple.
Scene #current / is-current-scene.
The apple / is-a / red.
\end{Verbatim}
\end{quote}

\begin{figure}
\begin{center}
\includegraphics[width=\columnwidth]{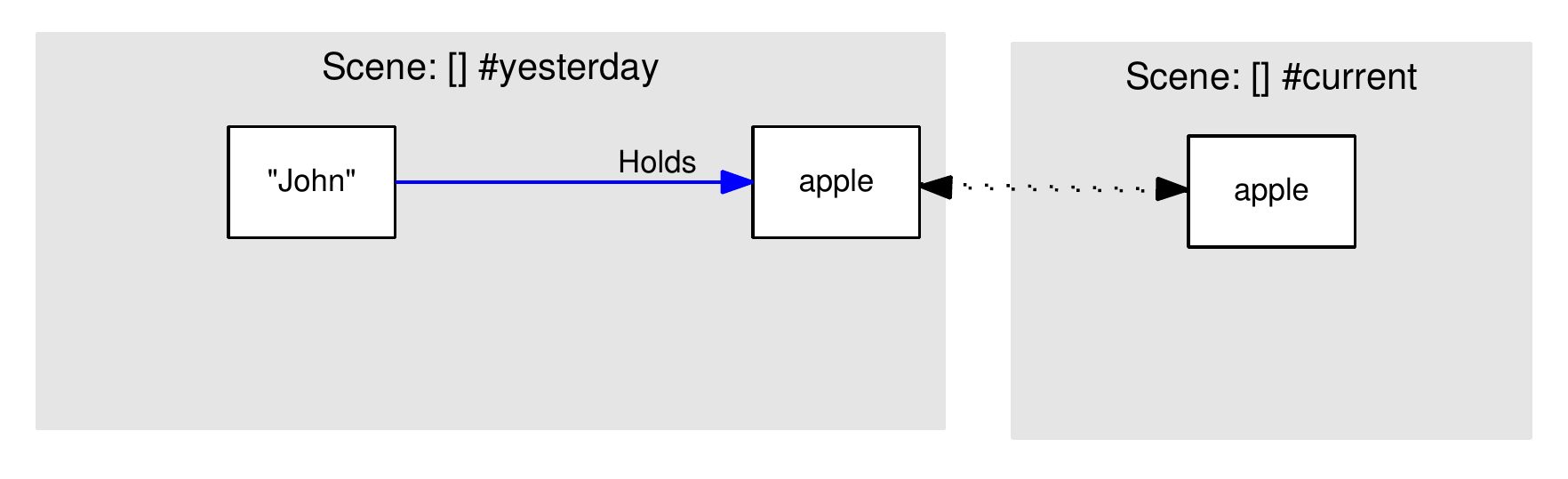} 
\caption{\label{fig:ComplexPhrases-801-FI} A noun modifier phrase refers to a different instance of the apple.
}
\end{center}
\end{figure}

The resulting instance structure in the focus is shown in Figure~\ref{fig:ComplexPhrases-801-FI}.

\section{Folk psychology}
\label{sec:FolkPsychology}

Folk psychology covers the way in which everyday people formalize the attribution of mental states to themselves and other humans, as well as the relationship between the humans' mental states and their past and future actions. Folk psychology is an important part of many stories, as some of them explicitly use the constructs of folk psychology as story elements. 

In this section we describe several examples of translating story snippets involving folk psychology from English to Xapi. 

%
%
\subsection{Understanding perceptions}

\begin{quote}{\em
I am watching Jay Leno and Brad Pitt talk on the television. Jay Leno says something, Brad Pitt laughs. I realize that Jay Leno was joking.
}
\end{quote}

There are several representational challenges here. First, there are two narrative spaces: the talk show, where Jay Leno and Brad Pitt are present, and my room, where (as far as this snippet goes) I am the only one present. This is naturally represented using two scenes in Xapagy, with the perception in scene \#Television appearing not directly, but through a quote-type construction through the character |Me| in the \#Reality scene. 

The other challenge is that the television perception and inference must be mixed up: events in the scene \#Television include things which the character |Me| sees, and some which he only infers. In the case of the joking of Jay Leno, we have a situation where the coincidence group had been formed from sentences which had been directly witnessed, and sentences which have been inferred. This is a very typical construct in Xapi stories.

Finally, the last challenge from the point of view of creating a Xapi text is that the interpretation was created not immediately after the witnessed action, but later. This means that we cannot use the simple version of the |thus| coincidence creation word. Instead we need to use a labeled version which allows us to refer back to a previous statement. Note that this kludge is not necessary if the interpretation is created internally through a headless shadow, which contains references to objects. 

The resulting Xapi code looks as follows:

\begin{quote}
\begin{Verbatim}
$NewSceneCurrent #Reality, none, man "Me"
$NewScene #Television, none, man "JayLeno", 
   man "BradPitt"
"Me" / sees in #Television // 
    "JayLeno" / talks-to #A / "BradPitt". 
"Me" / sees in #Television // 
    "BradPitt" / laughs.
"Me" / thinks in #Television // 
    "JayLeno" / thus #A jokes. 
\end{Verbatim}
\end{quote}

\begin{figure}
\begin{center}
\includegraphics[width=\columnwidth]{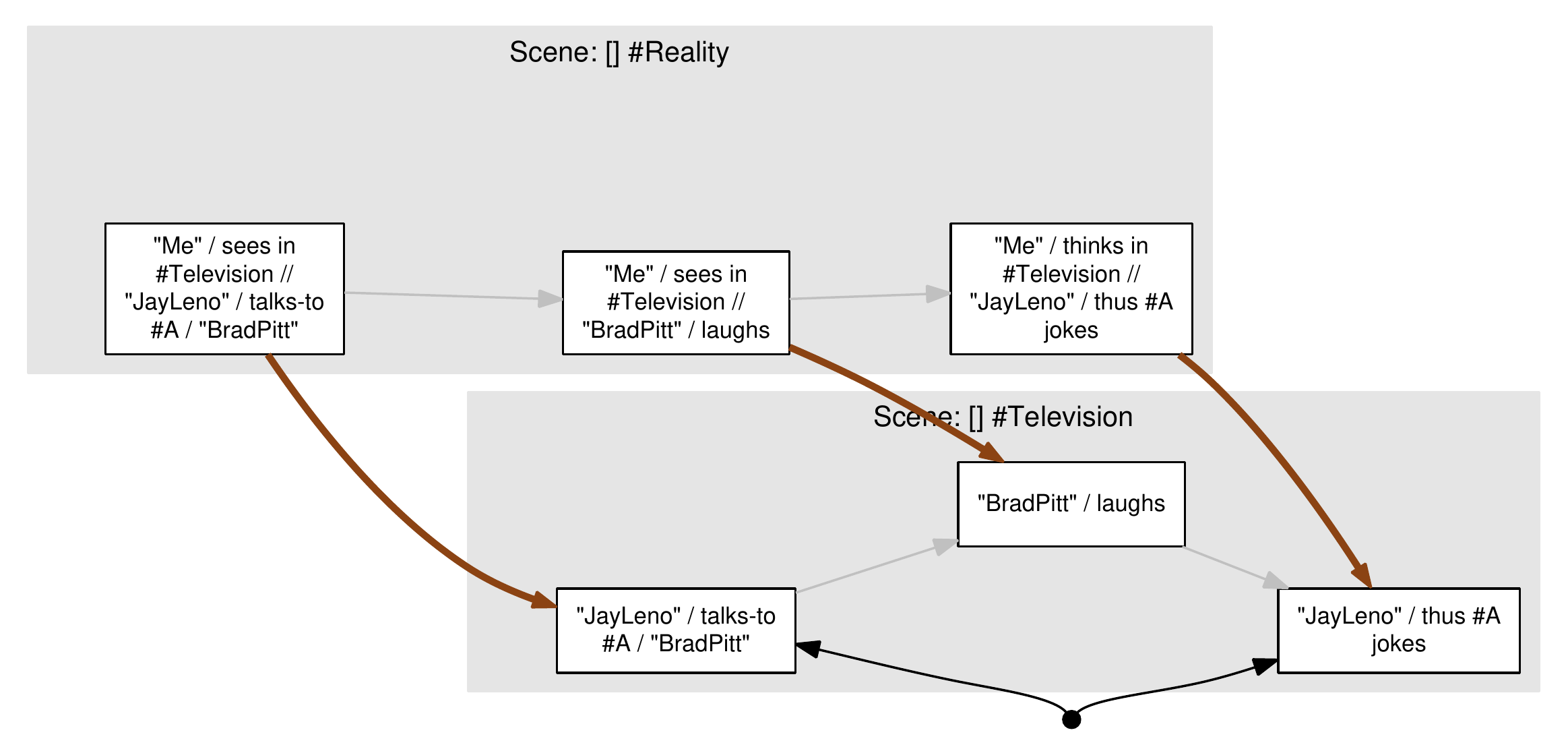} 
\caption{\label{fig:ComplexPhrases-401-FVI} The VIs in the focus after a series of perceptions where the interpretation is added to the perception later. 
}
\end{center}
\end{figure}

Parsing this Xapi story snippet creates a VI structure in the focus illustrated in Figure~\ref{fig:ComplexPhrases-401-FVI}. Note how the perception and thinking actions are in a natural succession of the statement. However, this succession is different in the \#Television scene where the interpretation snaps back to become coincidental with an earlier perception. 

As a note, such delayed interpretation can occur internally in Xapagy when a statement is created as a missing action. 

%
%
\subsection{Trying, succeeding and failing}

Human language frequently deploys constructs which show that an agent attempts to achieve a certain state of the world. Naturally, such story snippets must describe the given state of the world. Let us consider an example:

\begin{quote}{\em
George tries to put a cat in the box.
}
\end{quote}

The challenge in translating this statement to Xapi is that we need to describe a situation where the cat is in the box, which, however, is not necessarily true. The Xapagy model requires two scenes and {\em two instances of the cat and the box} to represent this. The following snippet implements this: 

\begin{quote}
\begin{Verbatim}
$NewSceneCurrent #Reality, none, man "John", 
   cat, box
$NewScene #Attempt, fictional-future, 
   man "John" -> man "John", 
   cat -> cat, box -> box
"John" / tries in #Attempt // 
    "John" / achieves.
"John" / tries in #Attempt // 
    The cat / thus is-inside / the box.
\end{Verbatim}
\end{quote}

\begin{figure*}
\begin{center}
\includegraphics[scale=0.7]{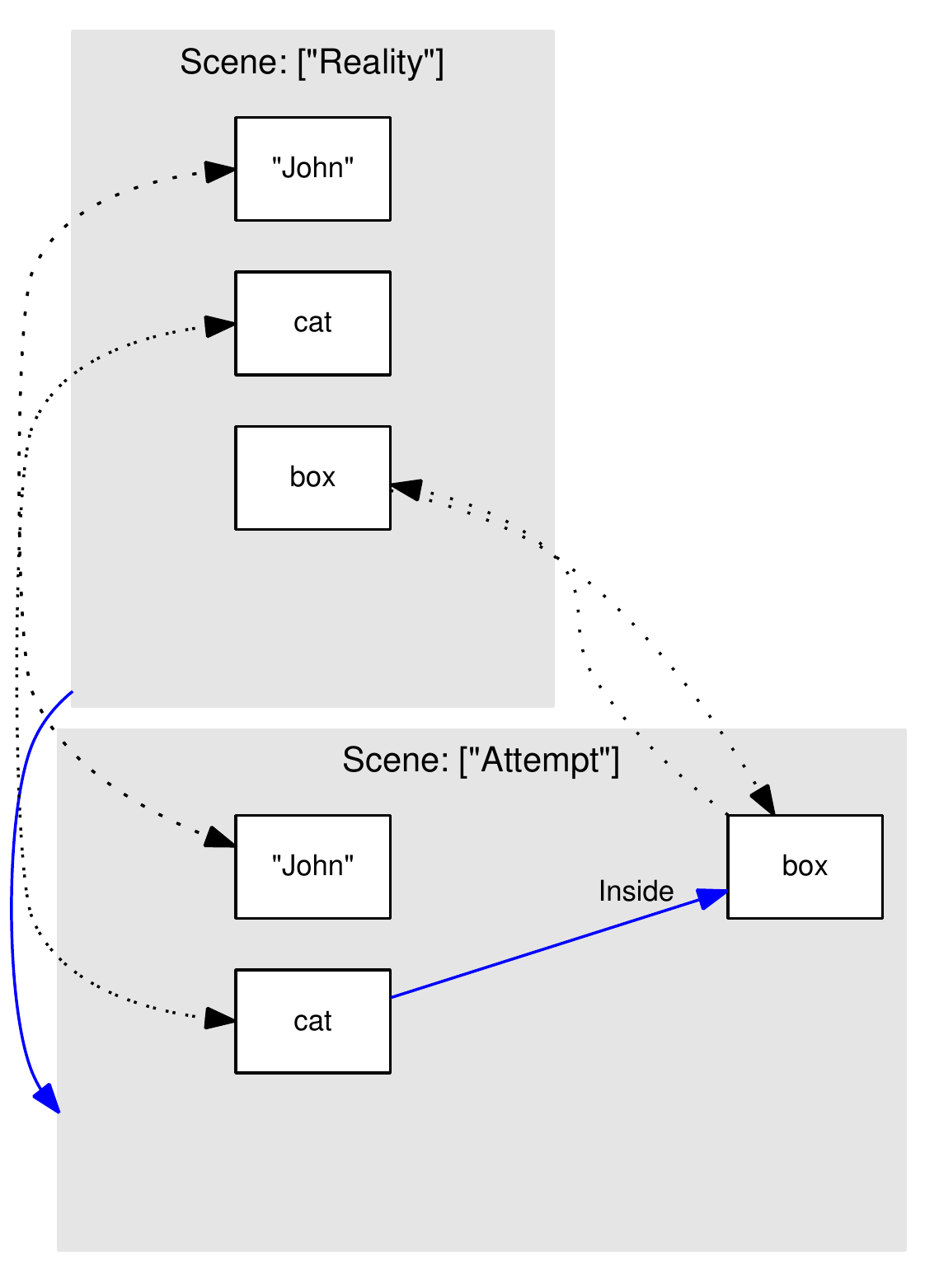} 
\caption{\label{fig:ComplexPhrases-301-FI} The focus instances after a snippet which describes an unsuccessful attempt to put a cat in the box. In the \#Attempt scene, we have a state of the world where the cat is in the box, in \#Reality, we don't.
}
\end{center}
\end{figure*}

The resulting focus instances can be seen in Figure~\ref{fig:ComplexPhrases-301-FI}.

While it goes beyond the scope of this technical report, let us quickly investigate what kind of predictions or inferences will Xapagy make in this situation? The HLS framework will recognize the specific ``trying'' structure in terms of achievement in a fictional future, and it can generate headless shadows for both success and failure. 

Success is simply a transposition of the achieves status in the fictional future into the current state:

\begin{quote}
\begin{Verbatim}
The cat / is-inside / the box.
\end{Verbatim}
\end{quote}

\noindent while failure is its negation:

\begin{quote}
\begin{Verbatim}
The cat / not-is-inside / the box.
\end{Verbatim}
\end{quote}

Finally, the code snippet above uses the specific Xapi L2 macro to create the attempt scene and set up the identities. Xapagy also has a specific meta-verb |clone-scene| which creates a cloned version of the current scene, sets up identity relations and a specific relation between the current scene and the newly created scene. 

\begin{quote}
\begin{Verbatim}
Scene / clone-scene / scene #Attempt.
\end{Verbatim}
\end{quote}

This is especially convenient for the inference and recall mechanism as it allows to implement a frequently used construct concisely. 

\if defined
{\em {\bf Future work: } The |clone-scene| meta-verb should also be able to take a relation as a parameter, such that the new scen will be connected to the current scene. } 
\fi

%
%
\subsection{Planning ahead}

In the previous example we discussed the case of an attempted action. A similar situation appears when an agent makes a possibly longer plan:

\begin{quote}{\em
John was hungry, so he decided to go to the kitchen, get some milk from the fridge and drink it.
}
\end{quote}

Before we move to the Xapi translation, let us first discuss what this natural language statement mean to a human observer. We can assume that John is not in the kitchen, he is engaged in some activity elsewhere. Although the statement describes a number of actions taking place, we know that these actions do not actually take place: for the time being, John will stay where he is. 

The Xapagy representational model for such situations is to create two instances of John. One instance of John, in the scene representing the current reality, will stay put and imagine the plan. The other instance of John, in the other scene, connected to the current one with a |future-fictional| relation is actually executing the steps necessary for the plan. The two instances of John are connected with an identity relation. The Xapi code for this sequence will be:

\begin{quote}
\begin{Verbatim}
$NewSceneOnly #Reality", none, man "John"
"John" / is-a / hungry.
$NewScene #Plan, fictional-future, 
    "John" -> "John"
"John" / plans in #Plan // 
    I / goes-to / a kitchen.
$..// I / open / a fridge.
$..// A milk / is-inside / the fridge.
$..// I / pick-up / the milk.
$..// I / drink / the milk.
\end{Verbatim}
\end{quote}

\begin{figure*}
\begin{center}
\includegraphics[scale=0.7]{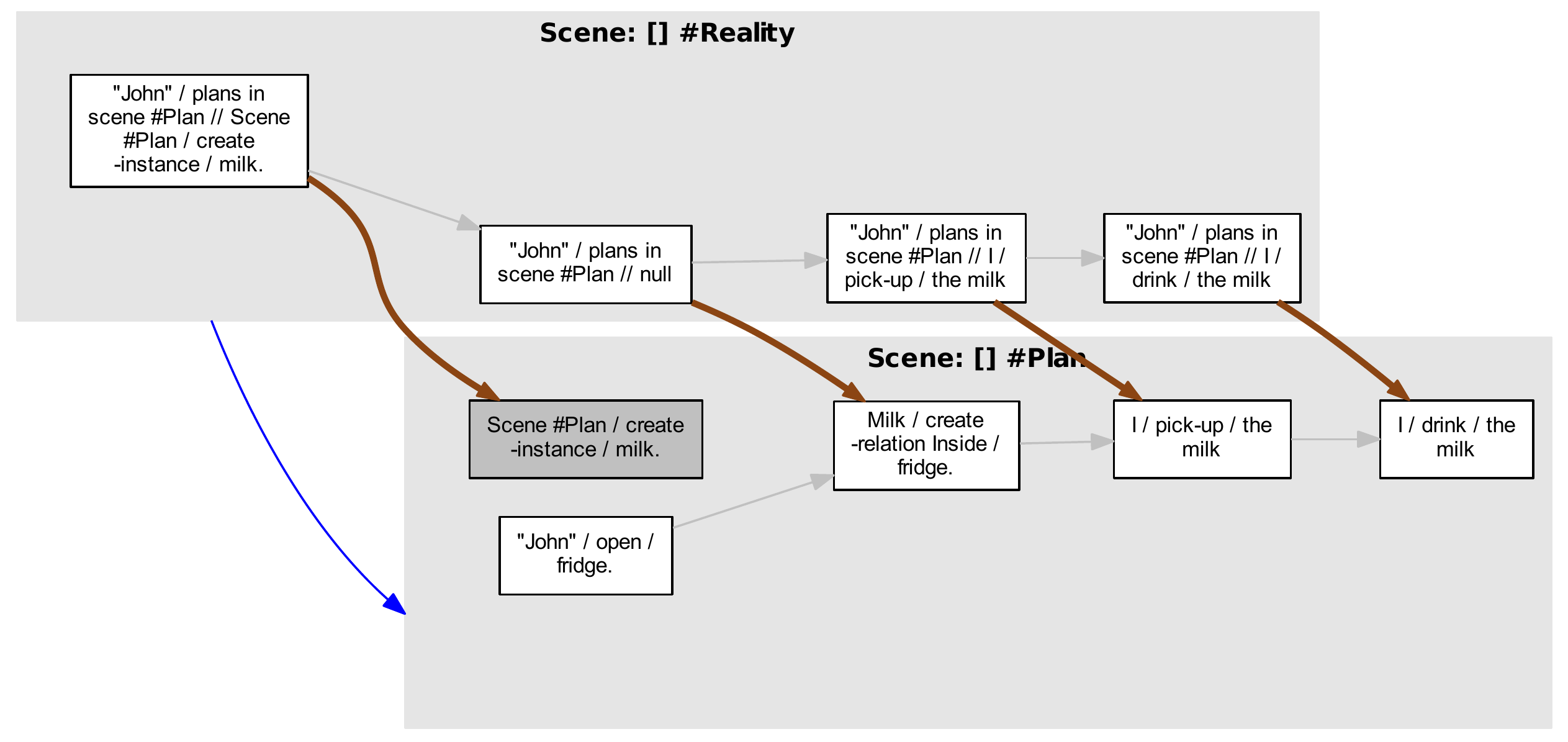} 
\caption{\label{fig:ComplexPhrases-201-FVI} The focus VIs after the creation of a plan. The plan is represented in a separate scene which describes the environment in which the plan is to take place. 
}
\end{center}
\end{figure*}

Executing this code will yield the instance and VI structures in Figure~\ref{fig:ComplexPhrases-201-FVI}.

As we are representing a plan, let us discuss what exactly makes this structure a plan. The first observation is that the scene \#Plan, by itself, does not have anything which would identify it as a plan (or even fictional). There is no ``fictional'' marker on the scene. It is only in relation to the original scene that the \#Plan scene is connected with the ``future-fictional'' relation. One implication of this is that we can represent multiple embeddings of scenes, each of them being fictional with relation to another. Also note that the scene \#Reality itself is not the reality: it is only an example in this paper. 
    
\if defined
{\em
{\bf Future work:} The current implementation has no model to reify the plan. We are talking about a series of intentions, but these intentions are not packaged into an instance which would be the plan, in the form of a series of intentions which can be written down. Thus, this model is not yet sufficient to represent sentences such as ``He had a plan'' or ``He changed his plans''. To allow such statements, we need to create an instance, which is then connected to the plan scene:

\begin{quote}
\begin{Verbatim}
He / has / a plan.
The plan / is-relation-content / scene "plan".
\end{Verbatim}
\end{quote}
}
\fi

%
%
\subsection{Enacting a plan}

Let us now discuss the representation of the enactment of the plan created in the previous section. Naturally, enacting a plan is just a series of actions. What we are interested in is how the enactment in the real scene will relate to the planning scene. The problem of how the plan drives the agent's concrete actions is an issue of story steering, and us such is beyond the scope of this paper. Nevertheless, even if we are looking at the agent from the outside, we still need to find the correspondences between his plan and its actions. To illustrate this internal view, let us discuss the representation of the following statement, which is a continuation of the one in the previous section:

\begin{quote}{\em
\ldots So John went to the kitchen and opened the fridge. The fridge was empty.
}
\end{quote}

The Xapi representation for this sentence (as following the previous ones) is as follows:

\begin{quote}
\begin{Verbatim}
"John" / goes-to / a kitchen.
The kitchen / is-identical / 
   the kitchen -- in -- "plan".
"John" / open / a fridge.
The fridge / is-identical / 
   the fridge -- in -- "plan".
The fridge / is-a / empty.
\end{Verbatim}
\end{quote}

\begin{figure*}
\begin{center}
\includegraphics[scale=0.7]{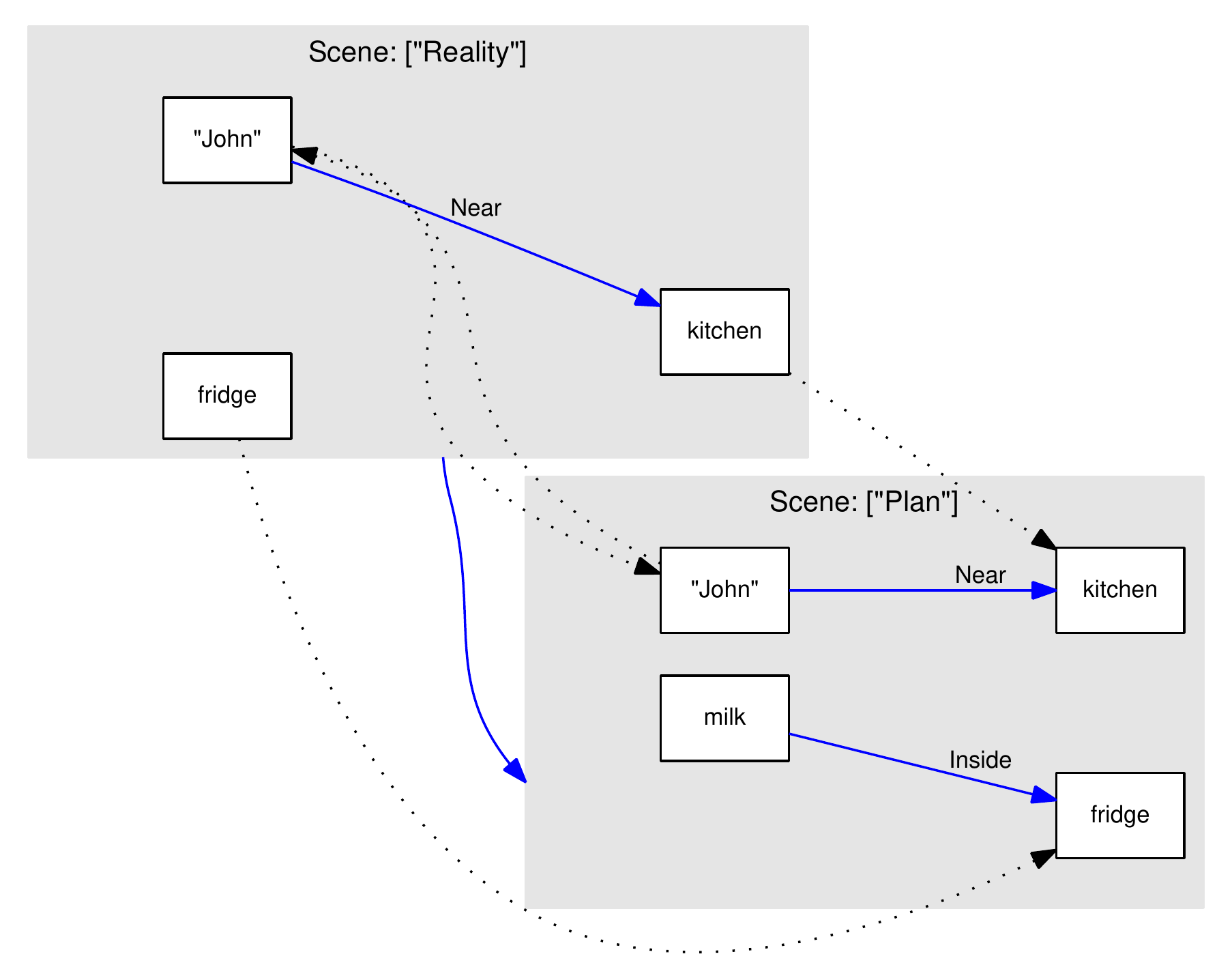} 
\caption{\label{fig:ComplexPhrases-202-FI} Mismatch between the plan and reality: there is no milk in the fridge.
}
\end{center}
\end{figure*} 

The relation between the plan and its execution is reflected in the identity relations: for instance, the kitchen into which John went is identity-related to the one in the plan. In the story snipped above, the identity relations are explicitly created in the story. In other circumstances, these relations can be inferred by the agent through a missing relation inference. It is also possible that a fast and careless reader (such as a Xapagy agent presented with a sequence of statements without interstitial pauses for the creation of headless shadows) would not even notice that what John is doing is actually an enactment of the plan.

\if defined
{\bf Discussion of subtleties:} We choose our natural language statement to be rather neutral here. Let us consider what other terminations we could have considered:

\begin{quote}{\em
\ldots but the fridge was empty.
}
\end{quote}

The word ``but'', as its presence actually refers to a plan which had failed. In particular we need a reification of the plan and a reification of the execution. Then, we can discuss about the failure of the execution attempt. 

\begin{quote}{\em
\ldots but there was no milk. 
}
\end{quote}

What is interesting here, is that the non existent milk in scene ``First'' must be linked back with an identity to the existing milk in scene ``plan''. 

These two situations will be discussed in future sections.  

\fi

\section{Unusual literary examples}
\label{sec:Unusual}

Let us now consider two more unusual examples of representation challenges, which illustrate the difficulty of translating complex expressions into Xapi. 

%
%

\subsection{Cicero disrespects Caesar: same action different meaning}


Greetings are formal exchanges between humans. While the greeting words normally have roots in comprehensible sentences, they are normally treated as opaque words. Let us consider the following exchange:

\begin{quote}
{\em
Cicero and Caesar met each other. Caesar said ``Ave!''. Cicero replied ``Salve!''.
}
\end{quote}

The first approximation for the Xapagy agent is to simply observe what just happened. This can be done by recording the utterances in an unparsed natural language form. 

\begin{quote}
\begin{Verbatim}
"Cicero" / meets / "Caesar".
"Caesar" / utters / text "Ave!".
"Cicero" / utters / text "Salve!".
\end{Verbatim}
\end{quote}

A next step would be to understand these two acts as greetings. For this, the agent does not need to perform a deep parse on the text: in fact, in the case of greetings, which are highly ritualized, conventional words, such deep parses are useless. The agent can recognize ``Ave'' and ``Salve'' as greeting words by simple matching. The interpretation of the act will be attached to the observed act through the coincidence relation. The interpretation, in this case, can be generate by the missing action mechanism. 

Note that the coincidence relation does not necessarily involve that the VIs have been created simultaneously, only that the VIs refer to different aspects of the same event. In this case, we have the observation and the interpretation of the event in the same coincidence group. As we shall see later, other combinations are also possible. The only restriction is that the VI must be still in the focus in order for the system to be able to add new VIs to its coincidence group.

The resulting code can be written in Xapi as follows:

\begin{quote}
\begin{Verbatim}
"Cicero" / meets / "Caesar".
"Cicero" / utters / text "Ave!".
"Cicero" / thus greets / "Caesar".
"Caesar" / utters / text "Salve!".
"Caesar" / thus greets / "Cicero".
\end{Verbatim}
\end{quote}

In addition to the interpretation of the individual utterances as greetings, we can also summarize the whole exchange as a single summary action:

\begin{quote}
\begin{Verbatim}
"Caesar" + "Cicero" / 
   in-summary meet-and-greet-each-other.
\end{Verbatim}
\end{quote}

\begin{figure*}
\begin{center}
\includegraphics[scale=0.5]{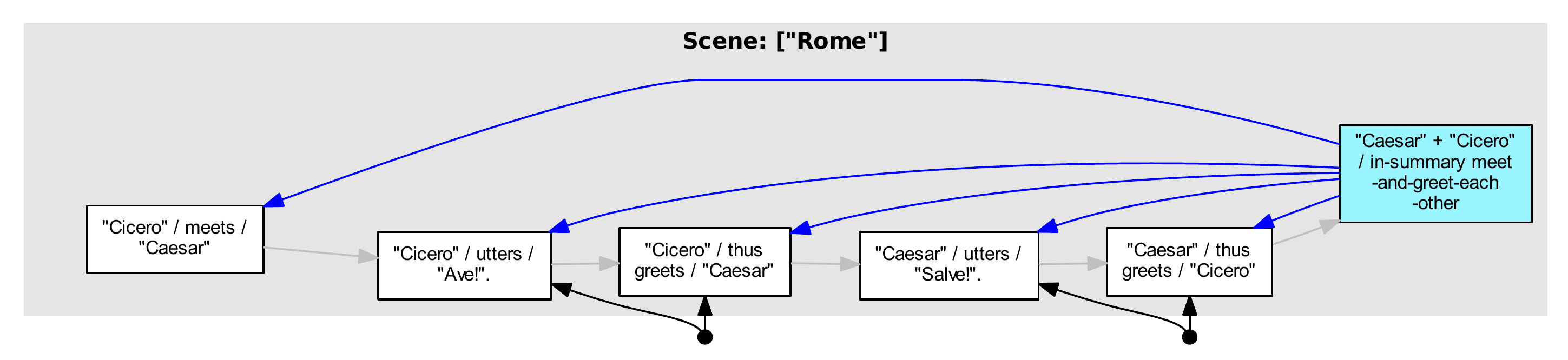} 
\caption{\label{fig:ComplexPhrases-003-FVI} The focus VIs for the example of Caesar and Cicero greeting each other.
}
\end{center}
\end{figure*}

Thus, the interpretation leads to the structure in Figure~\ref{fig:ComplexPhrases-003-FVI}.

%
%
\subsection{Lawrence of Arabia: evolving identity through recognition}

We have seen that changes in the attributes in real world objects are be represented by the change of instances. There are also situations when the real world entity does not change, but the  agent's changes its opinion about it (for instance, by correcting mistakes in attribute recognition). To illustrate such a case, let us consider the famous scene in Lawrence of Arabia, which is described on the website {\tt www.filmsite.org} as follows: 

\begin{quote}{\em
A dust cloud and then a tiny speck appear through shimmering, mirage-like heat waves on the desert horizon - Lawrence fears it is "Turks." The ominous image, more mirage than real, steadily enlarges and grows into a human being as it comes closer and closer.
}
\end{quote}

Let us now model this situation in Xapi. To avoid multiple quote sentences, we will write this from the point of view of Lawrence (to model it from the perspective of a third party observer, we can prefix each of these statements with |"Lawrence" / believes in scene|).

\begin{quote}
\begin{Verbatim}
A speck dust / is-far-away / "Me".
The speck dust / is-approaching / "Me".
The speck dust / changes / mirage.
The mirage / is-approaching / "Me".
The mirage / changes / a turk man + a turk man.
The turk man + the turk man / changes / a man.
The man / is-approaching / "Me".
The man / is-a / beduin.
The man / is-a / black-dressed.
\end{Verbatim}
\end{quote}

\begin{figure*}
\begin{center}
\includegraphics[scale=0.7]{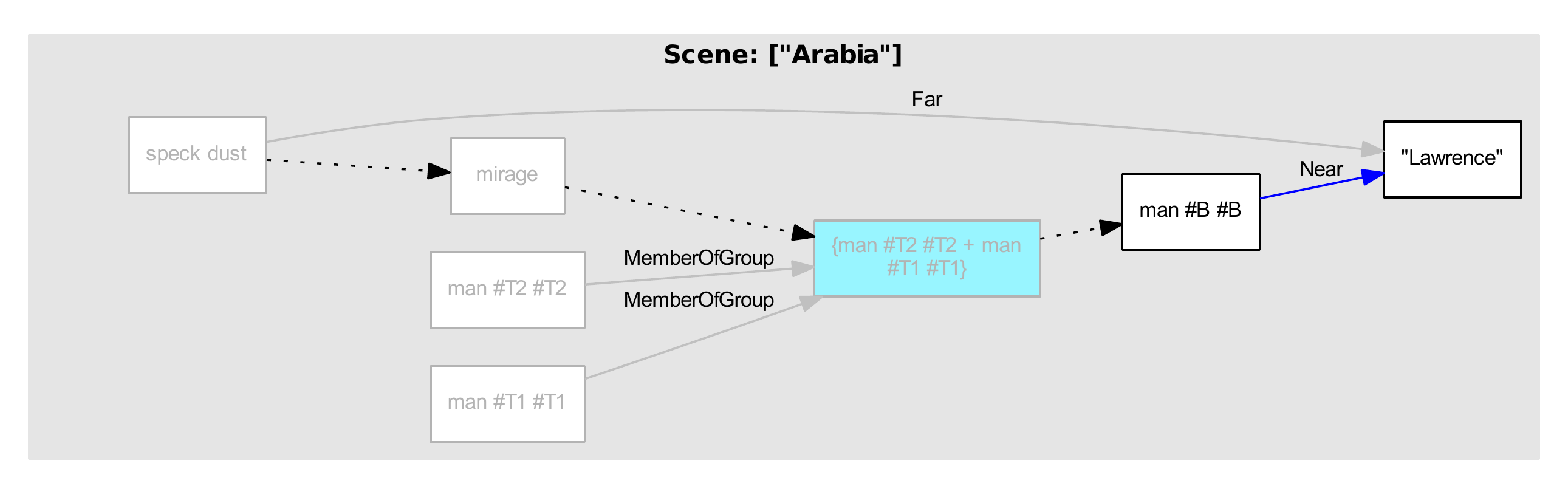} 
\caption{\label{fig:ComplexPhrases-101-FI} The instances in the current scene for he case of the beduin being recognized by Lawrence. The shaded instances are in focus, the unshaded ones are not.
}
\end{center}
\end{figure*}

In this case we have five changes of the instance which represent the entity to which Lawrence is looking. Note that the entity itself remains the same - there is a continuity of identity between the speck of dust first seen and the black-robed Beduin (Omar Sharif) who approaches and shots Lawrence's companion. This continuity is reflected through the chain of identities connecting the instances which represent the various interpretations: speck of dust, mirage, two Turks, one man. In fact, the number of instances created are even larger, because one of the interpretations actually refers to a group (two Turks). On the other hand, this example also illustrates that when a new discovery can be added as an additional attribute to the instance, we do not need to change the instance.

Now, of course, the Lawrence of Arabia scene is a famous example of successive revisions of incompatible  attributes. In real life, it is rare that the recognition process would proceed this slowly and through this many iterations. However, the gradual recognition of attributes (with or without the necessity of change) is an everyday occurrence.

\clearpage
\bibliography{../../Bibliography/Xapagy,../../Bibliography/ClassicStoryUnderstanding,../../Bibliography/CognitiveArchitectures,../../Bibliography/ComputationalNarrative,../../Bibliography/PhilosophyPsychology}
\bibliographystyle{aaai} 

\end{document}